\documentclass[runningheads]{llncs}

\usepackage{todonotes}
\usepackage[T1]{fontenc}
\usepackage{graphicx}
\usepackage{tcolorbox}
\usepackage[absolute,overlay]{textpos}
\usepackage{listings}
\usepackage{url}
\usepackage{hyperref}
\usepackage{subcaption}
\usepackage{booktabs}
\usepackage{lipsum}
\usepackage{siunitx}
\usepackage{multirow}
\usepackage{todonotes}

\author{\IEEEauthorblockN{Joshua Springer}
\IEEEauthorblockA{\textit{Department of Computer Science} \\
\textit{Reykjavik University}\\
Reykjavik, Iceland \\
\orcidlink{0000-0003-0137-1770} \href{https://orcid.org/0000-0003-0137-1770}{orcid.org/0000-0003-0137-1770}
}
\and
\IEEEauthorblockN{Gylfi Þór Guðmundsson}
\IEEEauthorblockA{\textit{Department of Computer Science} \\
\textit{Reykjavik University}\\
Reykjavik, Iceland \\
\orcidlink{0000-0003-0846-6617} \href{https://orcid.org/0000-0003-0846-6617}{orcid.org/0000-0003-0846-6617}}
\and
\IEEEauthorblockN{Marcel Kyas}
\IEEEauthorblockA{\textit{Department of Computer Science} \\
\textit{Reykjavik University}\\
Reykjavik, Iceland \\
\orcidlink{0000-0003-1018-3413} \href{https://orcid.org/0000-0003-1018-3413}{orcid.org/0000-0003-1018-3413}}
}

\AtBeginDocument{\nocite{*}}
\begin{document}

% \title{Toward A~Full Pipeline Approach to Autonomous Drone Landing Site Identification:\\From Terrain Survey to Embedded Classifier}
\title{Toward Appearance-based Autonomous Landing Site Identification for Multirotor Drones in Unstructured Environments}

\author{Joshua~Springer\orcidID{0000-0003-0137-1770} \and
Gylfi~Þór~Guðmundsson\orcidID{0000-0003-0846-6617} \and
Marcel~Kyas\orcidID{0000-0003-1018-3413}}

\institute{Reykjavik University, Iceland\\\email{\{joshua19, gylfig, marcel\}@ru.is}}

\authorrunning{Springer, Guðmundsson, Kyas.}
\titlerunning{Towards Autonomous Drone Landing Site Identification}

\maketitle

\begin{abstract}
A remaining challenge in multirotor drone flight is the
autonomous identification of viable landing sites in unstructured environments.
One approach to solve this problem is to create lightweight, appearance-based
terrain classifiers that can segment a drone's RGB images into safe and unsafe regions.
However, such classifiers require data sets of images and masks that can be prohibitively
expensive to create.
We propose a pipeline to automatically generate synthetic data sets
to train these classifiers,
leveraging modern drones' ability to survey terrain automatically
and the ability to automatically calculate landing safety masks
from terrain models derived from such surveys.
We then train a U-Net on the synthetic data set,
test it on real-world data for validation,
and demonstrate it on our drone platform in real-time.
\end{abstract}

\section{Introduction}
Autonomous multirotor drones are widely used in many fields,
primarily as remote sensor platforms.
While they excel at automated data collection, surveys, and other in-flight tasks,
autonomous landing remains a challenge in locations other than the initial takeoff site
or specially marked locations.
Furthermore, many drones are blind to obstacles and other hazardous terrain conditions,
as they primarily rely on GPS positioning at takeoff and landing.   
It is possible to increase landing accuracy and offer some obstacle avoidance by marking a safe landing site with a visual pattern, which the drone can locate via its (typically gimbal-mounted) RGB camera.
However, this requires that the landing site should be known beforehand
and that additional infrastructure -- possibly even requiring power -- should be in place.
To make landing in an unstructured, previously unknown environment viable,
the drone must first analyze its environment in flight using more complex sensors, e.g., LiDAR
or stereo depth cameras.
Such advanced sensors are effective but also
computationally expensive and thus power-hungry, reducing task-oriented mission time.
Alternatively, the heavy computation can be offloaded to a ground station,
but this requires added infrastructure
and introduces problems with transmission overhead, connectivity, and latency.

We approach autonomous landing site identification as an image segmentation problem,
with the goal of creating an appearance-based classifier that distinguishes safe and unsafe landing
sites in images from our drone's gimbal-mounted RGB camera.
Image segmentation often requires a labeled data set,
which is potentially prohibitively expensive to create manually.
We can exploit drones' strength in surveying terrain to easily build terrain models from which we
can automatically generate a synthetic, labeled data set to circumvent this challenge.
We prioritize keeping the computational overhead as low as possible as we run our solution onboard a drone
in real time.
%, without significantly reducing the in-flight mission time.
We target the typical, gimbal-mounted RGB camera as our primary sensor instead of LiDAR,
as
the RGB camera is the most common drone peripheral sensor and will make the solution more generalizable and lightweight.
Our contributions are as follows:
(1)~we present a pipeline to create synthetic image segmentation data sets
for determining landing safety from terrain surveys
(see Sections~\ref{section:3d_terrain_model_reconstruction}
and~\ref{section:segmentation_dataset_generation}).
(2)~We present a method for creating small, but effective, real-world validation sets of videos taken of
known-safe and known-unsafe landing sites in the real world to determine whether our method can bridge
the gap from simulation to reality (see Section~\ref{section:validation_dataset}).
Finally, (3)~we showcase 
%conduct a demonstration of 
a U-Net trained on one such data set that is able to 
correctly classify 15 of 18 validation cases, which run in real-time onboard our drone platform.

We evaluate our work using 
%test on 
the drone platform described in~\cite{cbmi_demo_payload_paper},
which has previously been demonstrated successfully in the context of autonomous landing
with an RGB camera and visual markers~\cite{visir_landing}.
We have added a Google Coral TPU accelerator to run our tiny terrain classifier
(under 1 MB in size).
While the small classifiers are best suited to specific environment types,
they are small enough that we can store multiple classifiers on
an embedded computer that can use the most relevant one.
On the other hand, the method for creating the classifiers is generalizable to any environment
where there is some relationship between visual appearance and landing safety.
The final stage of this process is the actual autonomous landing of the drone,
which is out of the scope of this paper.
Our approach of identifying a viable landing site in RGB video as an image segmentation problem,
where we mark each pixel as safe or unsafe,
will allow our system to choose a safe pixel position representing a target landing location.
With minor future adaptations,
our system will then be able to control the drone and carry out landings
using the method described in~\cite{visir_landing},
which lands the drone at a particular site specified by a pixel position.

\section{Related Work}
\label{section:related_work}

Landing site detection is similar to the notion of \emph{traversability},
which has been explored extensively on ground vehicles.
\cite{survey_ground_vehicles_traversability}
presents a survey of ground vehicle traversability methods,
dividing them generally into appearance-based, geometric-based, and mixed methods.
Appearance-based methods analyze terrain using visual sensors,
whereas geometric-based methods require sensors such as LiDAR or RGBD
that can extract a 3D representation of the environment.
Some methods show success in traversability analysis, where training data is collected
and labeled directly with 3D sensors,
and terrain classification is performed with visual sensors 
only~\cite{weakly_supervised_path_segmentation_icra,unsupervised_traversability_offroad}.
Some others classify 3D data from LiDAR and RGBD cameras directly~\cite{automatic_3d_pointcloud_labeling_gazebo}
There is also a tendency to classify each pixel into one of three groups, e.g.,
traversable, not traversable, and unknown.
These methods typically do not analyze reconstructed terrain models,
but instead, analyze the terrain directly via some sensor
and apply that analysis to label RGB images.
Many traversability methods are developed and tested in simulation,
but the gap from simulation to reality is often not overcome.
% GTG: This sentence is a bit confusing
While appearance-based methods can lack accuracy, they only require simple RGB camera sensors
that are abundant, cheaper, and have a lower processing overhead than the hardware needed for 
%simpler sensors, e.g., typical RGB cameras, that are less expensive both to procure and in terms of data processing requirements, than 
geometric-based methods~\cite{survey_ground_vehicles_traversability}.
Many of the methods presented require specialized data sets with multiple data sources, i.e., RGB and at least one LiDAR or RGBD, to determine a label for the terrain.

Most work in autonomous landing site identification in unstructured environments is geometric-based, requiring
real-time LiDAR or RGBD analysis, full-size GPUs, etc.~\cite{uav_computer_vision_survey,lidar_automatic_rgb_manual}.
However, some methods visually locate their starting location using its visual appearance only~\cite{scherer_landing_visual_teach_repeat_fisheye}.
More general appearance-based methods in this context are less widespread,
as they often depend on expensive, manually-labeled data sets.
In this context, some methods exist to 
sparsely label video frames manually, and then propagate the labels from frame to frame~\cite{iterative_label_propagation};
others use previously trained neural networks to add new labels to drone 
videos~\cite{yolo_object_avoidance_okutama_simulation}.
These methods have not yet been applied to autonomous landing but have potential in identifying
viable landing sites.
Many methods are tested only in simulation or laboratory settings on real drone video but are not embedded onto the drone~\cite{airsim_multistage_xavier,yolo_object_avoidance_okutama_simulation}.
SafeUAV serves as an initial proof of concept for our proposed method~\cite{safeuav}.
It uses existing synthetic data sets from Google,
making it possible to quickly extract training data from many locations
but making the data less dynamic when the environment has recently changed.
Their setup requires a camera at a fixed angle of $45^\circ$ below the horizontal,
and their classifiers predict the scene's 3D depth and landing safety.
Crucially, they consider the problem of embedding their classifiers on a drone in the real world
and therefore test them on embeddable hardware (an NVIDIA Jetson TX2) and actual drone footage,
although they do not deploy their solution on a drone.
% It also allows for less customizability in the parameters of the terrain reconstruction.

We take inspiration from many of these papers and seek to create a full-pipeline approach that allows for flexibility
in data requirements, minimizes manual labeling as much as possible, and ultimately produces a viable embedded terrain classifier that can run in real-time onboard a drone.
To satisfy the real-time aspect,
we prefer to use an appearance-based approach such that the classifier can perform
inference on RGB images, which are lightweight to analyze when compared to, e.g., LiDAR data.
This also makes the method easier to generalize to drone platforms without LiDAR or RGBD cameras.
For flexibility in data requirements and ease in manual labeling, we generate intermediate 3D models
from which we produce a synthetic data set of RGB images and masks.
This gives the advantage that we can use many different data sources to generate the intermediate models, e.g., photogrammetry, LiDAR, RGBD, etc.
This also means that we do not necessarily have to collect our data
but can use openly available, standard-format, unlabeled data sets from terrain surveys.
We also allow for the ability to quickly add manual labels to the intermediate models one time;
such labels then propagate to all of the images generated.
Importantly, we can vary the angle of our camera,
which inherently makes the classifier more flexible than that in~\cite{safeuav}.
Finally, while we do not prescribe a particular, optimal classifier architecture,
we create a successful U-Net that is relatively tiny (on the order of 1 MB instead of more than 1 GB)
and can be deployed on power-efficient hardware compared to all methods described earlier.
We also showcase how our classifier can be embedded onboard a drone.

\section{Methods}
We describe the process of collecting and transforming terrain data for our image segmentation purposes.
We have automated this process
except for logistical tasks such as transporting the drone to the survey location and optional, manual label refinement.

\begin{figure}[tb]
    \centering
    \includegraphics[width=0.95\linewidth]{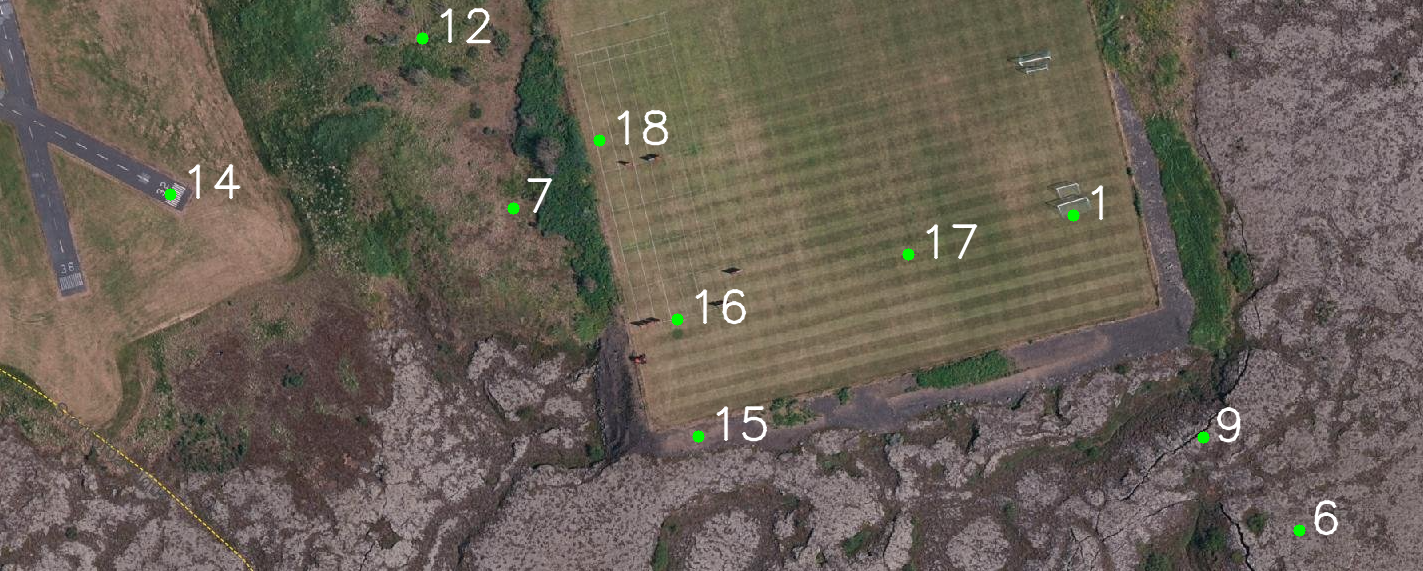}
    \caption{One of three data collection sites. We manually and randomly picked 18 validation sites, and we number them as follows, marking safe with an S and unsafe with a U:
(1U) -- an archery target on a soccer field,
(2U) -- a bush,
(3S) -- a flat, dirt area,
(4U) -- a large, cracked rock mound,
(5U) -- high vegetation area,
(6S) -- flat, mossy area in a lava field,
(7S) -- flat, grassy area,
(8S) -- flat, mossy area in a lava field,
(9U) -- crack in a lava field,
(10S) -- dirt patch in a lava field,
(11S) -- road,
(12U) -- person,
(13U) -- very rough lava field,
(14S) -- model aircraft runway,
(15U) -- sloped, gravel edge of a soccer field,
(16S) -- green spot in a soccer field,
(17S) -- middle of a soccer field,
(18U) -- soccer goal.
Map source: Loftmyndir ehf.~\cite{loftmyndir}
}
    \label{figure:data_collection_sites}

\vspace{-0.1cm}
\end{figure}

\subsection{Data Acquisition -- Terrain Surveys}
\label{section:terrain_surveying}
The first step in generating data for the image segmentation problem is to collect terrain data from an
environment similar to where a drone will need to land autonomously.
For this purpose, we test both photogrammetry and LiDAR surveys,
which have been automated in the case of many drones so that an operator needs only to select an area for surveying
and then can deploy the drone to do the survey automatically.
Photogrammetry is the process of combining images to create composite images or 3D models.
We use ortho and oblique images in photogrammetry because this results in better 3D reconstructions of surfaces of all orientations.
For example, when using primarily orthogonal images for terrain reconstruction, vertical surfaces can be severely
distorted, as mentioned in~\cite{safeuav}.
LiDAR surveys provide similar terrain reconstruction capabilities as photogrammetry, but often with higher quality,
higher data collection speed, and a higher price point -- although this is not a hard rule.
LiDAR produces point cloud data, which can be colorized by registering the points with RGB images collected simultaneously.

\subsection{Data Acquisition -- Validation Dataset}
\label{section:validation_dataset}

To generate a validation data set without frame-by-frame, manual labeling,
we collect 10-second videos of particular validation sites in the field.
In each video, the drone's camera holds the validation site in the center of the frame,
such that it collects many different frames from many different angles
as the drone moves.
The camera's tilt is between $45^\circ$ below the horizon and vertically down.
We set the validation site to be either an obstacle or a clearing and record the classifier's predictions
of the center pixels of the video for all the frames, adjusting the central region's size according to the validation site's size.
We tag each video as a whole according to whether it shows a safe landing site or not,
and we compare our manual classification to
the classifier's predictions over the frames of the video.
The prediction describes the classification of the majority of pixels in the central region
of the video, and we aggregate it over time to determine a simple, binary prediction for the
entire video.
Some of the sites are hand-selected over a range of anticipated difficulties,
considering the appearance-based classifier's lack of geometric understanding of the scene it is classifying.
For example, we expect the classifier to easily determine that an open field is safe
and that a large crack in the ground is unsafe.
On the other hand, we expect that it should be hard to classify slanted dirt areas as unsafe,
since they appear visually similar to the safe, level dirt areas.

\subsection{3D Terrain Model Reconstruction}
\label{section:3d_terrain_model_reconstruction}

The second step in generating our dataset,
after conducting a survey or downloading an openly available survey,
is to generate an RGB mesh that is a colorized, 3D depiction of the terrain.
We create this mesh from photogrammetry data using WebODM~\cite{webodm}, 
or from LiDAR data by performing a Poisson reconstruction~\cite{poisson_surface_reconstruction}
in Cloudcompare~\cite{cloudcompare}.

\begin{figure}[ht]
        \centering

        \begin{subfigure}[t]{0.32\linewidth}
                \includegraphics[width=\linewidth]{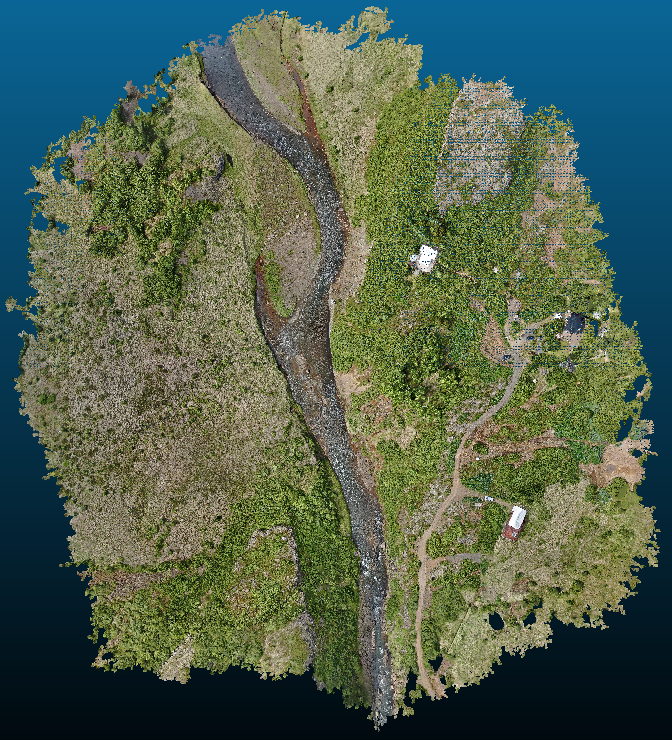}
                \caption{RGB, unsegmented}
        \end{subfigure}
        \begin{subfigure}[t]{0.32\linewidth}
                \includegraphics[width=\linewidth]{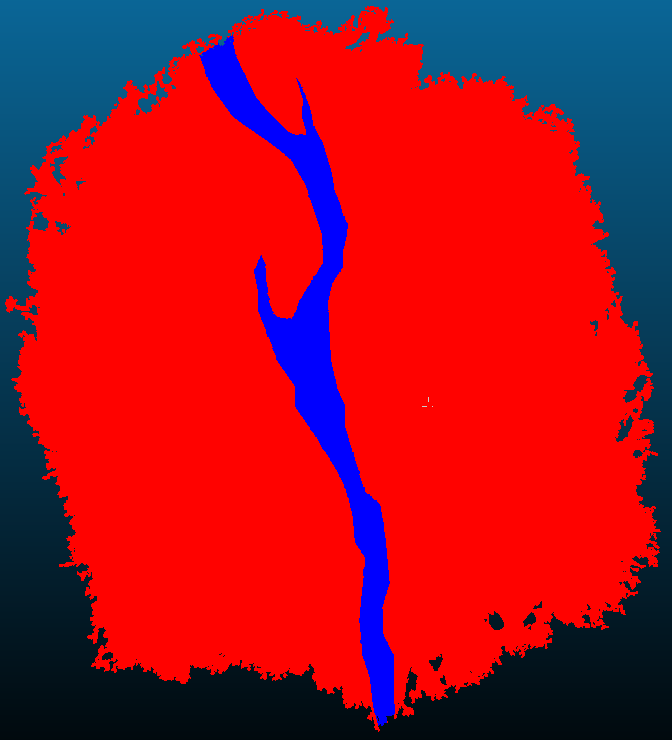}
                \caption{Manually segmented}
        \end{subfigure}
        \begin{subfigure}[t]{0.32\linewidth}
                \includegraphics[width=\linewidth]{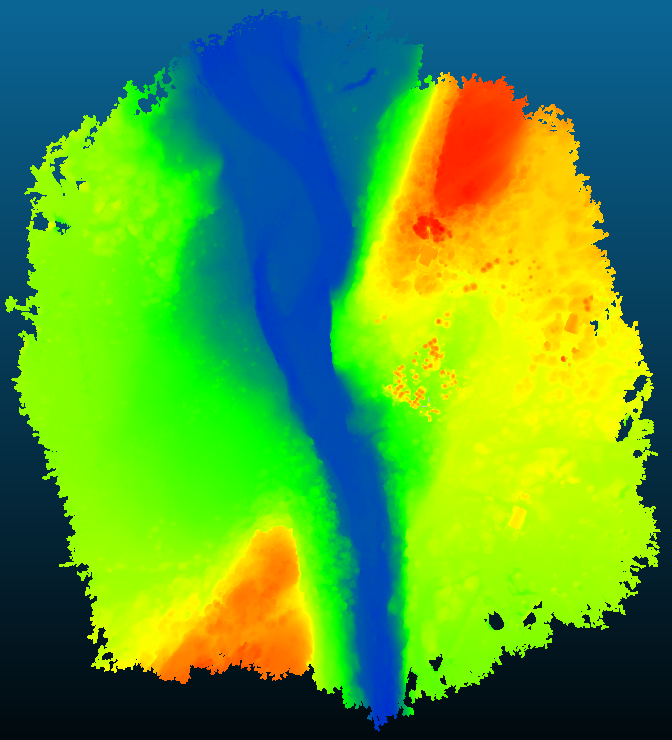}
                \caption{$Z$-coordinate}
        \end{subfigure}
        
        \caption
        {
            Example of manual segmentation of a river in the summer house dataset by isolating the river and adding
            a manual classification to it in CloudCompare.
            It is not feasible to isolate the river by simply filtering on the altitude above sea level (ASL)
            since the terrain has a significant slope.
        }
        \label{figure:isolating_river_summer_house_11_july_2024}
\end{figure}

The third step is to create a ``label mesh'' with the same topography as the RGB mesh
that is marked according to which regions are safe and unsafe according to our geometric specifications.
We sample the RGB mesh to create a point cloud with uniform density to calculate geometric features.
This is necessary to remove ``fuzziness'' in point clouds derived from photogrammetry,
and to remove scanning overlap in point clouds collected via LiDAR,
since such variations can influence geometric features.
We then calculate the normal vectors and geometric features of
\emph{verticality} -- how slanted a region is --
and \emph{surface variation} -- how rough a region is~\cite{contour_detection_unstructured_3d_point_clouds}.
We compute a binary safety metric over the mesh, marking as unsafe all areas with
verticality $> 0.01$,
surface variation $> 0.002$ (experimentally determined).
We further eliminate unsafe areas that have been classified as safe
geometrically, because of their low verticality and surface variation,
e.g., lakes and rivers, by simply selecting them in CloudCompare and adding a manual classification to all the points
representing the problematic surface.
This process is quick, requiring only about 2 minutes to manually label the river of several hundred meters in
Figure~\ref{figure:isolating_river_summer_house_11_july_2024}.
We then apply Gaussian smoothing to the safety metric as the new grayscale texture for the label mesh.
Finally, we slice the meshes into smaller chunks that are manageable on our hardware.

\subsection{Synthetic Segmentation Dataset Generation}
\label{section:segmentation_dataset_generation}

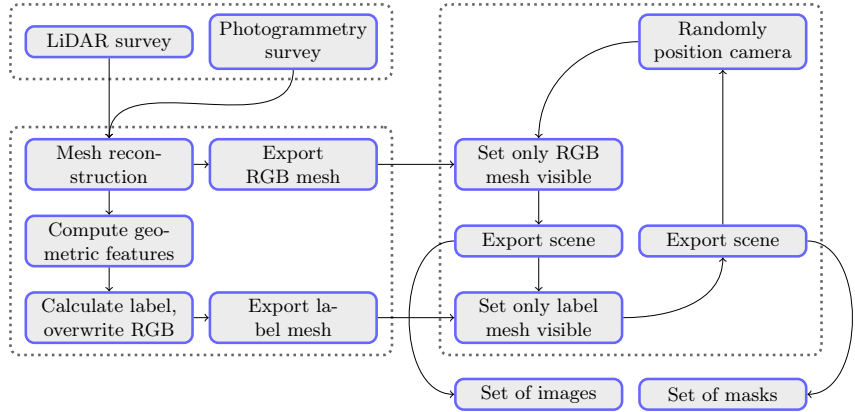
\begin{figure}[tb]
    \centering
    \resizebox{0.95\textwidth}{!}{
    \begin{tikzpicture}[scale=1,
        roundnode/.style={circle, draw=green!60, fill=gray!5, very thick, minimum size=7mm},
        squarednode/.style={rounded corners, draw=black!60, dotted, fill=gray!0, very thick, minimum size=5mm},
        blacksquarednode/.style={rounded corners, draw=black!60, fill=gray!15, very thick, minimum size=5mm},
        redsquarednode/.style={rounded corners, draw=red!60, fill=gray!15, very thick, minimum size=5mm},
        bluesquarednode/.style={rounded corners, draw=blue!60, fill=gray!15, very thick, minimum size=5mm},
        ]
        
        \node[squarednode, text width=6cm, text height=1cm, align=center] (data_collection) at (0,2) { };
        \node[squarednode, text width=6cm, text height=3.5cm, align=center] (data_collection) at (0,-1.25) { };
        \node[squarednode, text width=6cm, text height=5.5cm, align=center] (data_collection) at (7,-0.25) { };
        
        \node[bluesquarednode, text width=2.5cm, align=center] (lidar) at (-1.5,2) { LiDAR survey };
        \node[bluesquarednode, text width=2.5cm, align=center] (photogrammetry) at (1.5,2) { Photogrammetry survey };
        
        \node[bluesquarednode, text width=2.5cm, align=center] (reconstruction) at (-1.5,0) { Mesh reconstruction };
        \node[bluesquarednode, text width=2.5cm, align=center] (export_rgb) at (1.5,0) { Export RGB mesh };
        \node[bluesquarednode, text width=2.5cm, align=center] (geometric_features) at (-1.5,-1.25) { Compute geometric features };
        \node[bluesquarednode, text width=2.5cm, align=center] (label_generation) at (-1.5,-2.5) { Calculate label, overwrite RGB };
        \node[bluesquarednode, text width=2.5cm, align=center] (export_label) at (1.5,-2.5) { Export label mesh };
        
        \draw[->] (lidar) to [out=-90,in=90] (reconstruction);
        \draw[->] (photogrammetry) to [out=-90,in=90] (reconstruction);
        \draw[->] (reconstruction) to [out=-90,in=90] (geometric_features);
        \draw[->] (geometric_features) to [out=-90,in=90] (label_generation);
        
        \draw[->] (reconstruction) to [out=0,in=180] (export_rgb);
        \draw[->] (label_generation) to [out=0,in=180] (export_label);
        
        \node[bluesquarednode, text width=2.5cm, align=center] (randomly_position_camera) at (8.5,2) { Randomly position camera };
        \node[bluesquarednode, text width=2.5cm, align=center] (rgb_mesh_visible) at (5.5,0) { Set only RGB mesh visible };
        \node[bluesquarednode, text width=2.5cm, align=center] (export_picture_1) at (5.5,-1.25) { Export scene };
        \node[bluesquarednode, text width=2.5cm, align=center] (label_mesh_visible) at (5.5,-2.5) { Set only label mesh visible };
        \node[bluesquarednode, text width=2.5cm, align=center] (export_picture_2) at (8.5,-1.25) { Export scene };
        
        \draw[->] (export_rgb) to [out=0,in=180,style=dotted] (rgb_mesh_visible);
        \draw[->] (export_label) to [out=0,in=180,style=dotted] (label_mesh_visible);
        
        \draw[->] (randomly_position_camera) to [out=-180,in=90,style=dotted] (rgb_mesh_visible);
        \draw[->] (rgb_mesh_visible) to [out=-90,in=90,style=dotted] (export_picture_1);
        \draw[->] (export_picture_1) to [out=-90,in=90,style=dotted] (label_mesh_visible);
        \draw[->] (label_mesh_visible) to [out=0,in=-90,style=dotted] (export_picture_2);
        
        \draw[->] (export_picture_2) to [out=90,in=-90,style=dotted] (randomly_position_camera);
        
        \node[bluesquarednode, text width=2.5cm, align=center] (rgbs) at (5.5,-3.75) { Set of images };
        \node[bluesquarednode, text width=2.5cm, align=center] (masks) at (8.5,-3.75) { Set of masks };
        
        \draw[->] (export_picture_1) to [out=180,in=180] (rgbs);
        \draw[->] (export_picture_2) to [out=0,in=0] (masks);
    \end{tikzpicture}
    }
    \caption{Pipeline for creating labeled image datasets from terrain surveys.}
    \label{figure:dataset_creation_pipeline}
\end{figure}

\begin{figure}[tb]
    \centering
    \includegraphics[width=0.32\linewidth]{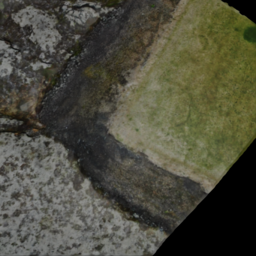}
    \includegraphics[width=0.32\linewidth]{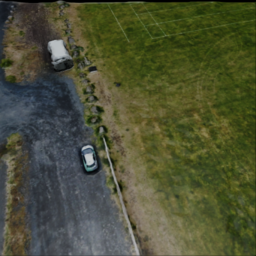}
    \includegraphics[width=0.32\linewidth]{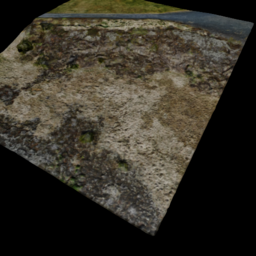}\\
    \includegraphics[width=0.32\linewidth]{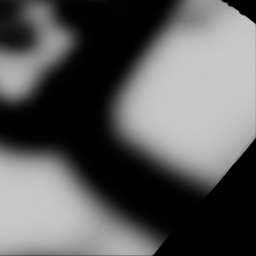}
    \includegraphics[width=0.32\linewidth]{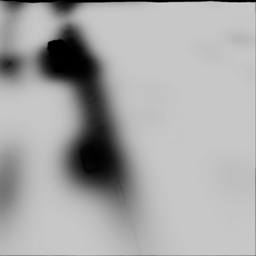}
    \includegraphics[width=0.32\linewidth]{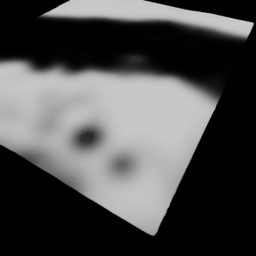}
    
    \caption{Example images and masks from the synthetic data set.}
    \label{figure:dataset_example}
\end{figure}

The fourth and last step in generating our data set is to create synthetic aerial images representing
a drone's view of the terrain and corresponding masks that specify which pixels in those images represent safe landing sites.
Using NVIDIA Isaac Sim~\cite{nvidia_isaac_sim},
we create a scene with the RGB and label meshes at the same position and orientation,
with only one visible at a time.
We position a virtual camera randomly in the scene,
and aim it at a random location on the meshes,
ensuring that the camera is between a minimum and maximum height
above the terrain and between a minimum and maximum angular deflection from vertical down.
We set the RGB mesh as visible and take a picture, effectively creating a typical aerial picture of a given terrain area.
Then, we set the label mesh as visible and take another picture, creating a safety mask for the RGB image.
We repeat this process for each slice to generate the labeled data set as many times as necessary.
Figure~\ref{figure:dataset_creation_pipeline} visualizes this process and 
Figure~\ref{figure:dataset_example} shows examples from a dataset, where the terrain is shown on top,
and masks are shown on the bottom.
White and black areas indicate safe and unsafe areas for landing, respectively.

The following is a non-exhaustive list of changeable parameters which we have set experimentally
and which may require special attention depending
on the specific scenario:
data collection altitude, image/pointcloud overlap, 
method and radius for calculating point cloud normal vectors,
Poisson reconstruction parameters,
radii for calculating verticality and planarity (as well as the corresponding thresholds),
slice size,
maximum height and deflection of the virtual camera.
These will significantly affect the quality of the datasets generated,
and they may vary case by case.

\subsection{Terrain Classifiers}
\label{section:Terrain_Classifiers}

\begin{table}[tb]
    \caption{Datasets used to train the terrain classifiers. We use both LiDAR and photogrammetry (p-gram.) data.}
    \label{table:datasets}
    \centering
    \begin{tabular}{llllrl}
    \toprule
    {} & Location & Type & Sensor & Data points & Source \\
    \midrule
    1 & RC airfield & p-gram. (oblique) & DJI H20T  & 750 & own \\
    2 & RC airfield & LiDAR & DJI L2 & 750 & own \\
    % 3 & Sandkey & ortho photogrammetry & unknown & WebODM \\
    3 & Sheffield Cross & p-gram. (ortho) & unknown & 241 & WebODM \\
    4 & Soccer field & LiDAR & DJI L2 & 750 & own\\
    5 & Soccer field & p-gram. (oblique) & DJI H20T & 750 & own \\
    6 & Summer house & p-gram. (oblique) & DJI H20T & 750 & own \\
    \bottomrule
    \end{tabular}
\end{table}

We use deep learning methods to generate classifiers for our image segmentation dataset described in Section~\ref{section:segmentation_dataset_generation}.
Although this is a typical segmentation problem, and there are many deep learning frameworks we could use,
we are targeting a specific inference platform: the Google Coral TPU~\cite{google_coral_edge_tpu},
which imposes a particular toolchain revolving around Tensorflow~\cite{tensorflow,tensorflowlite},
as well as a particular set of available operations, input sizes, etc.

We test the U-Net architecture for this task based on its well-known strong performance
in segmentation problems. %: U-Net, fully convolutional, and SegNet.
However, this is a non-exhaustive list, and other architectures could be applicable,
e.g., auto-encoders, fully convolutional, and others.
Additionally, initial experimentation revealed a few key notions.
First, the little available memory on the Google Coral imposed size limitations of about 8 MB on our models
and the input image resolution.
Second, longer chains of operations, e.g., more sections in the U-Net architecture, significantly affected the framerate
at which the Google Coral could perform inference, even when all of the overhead operations, such as playing/resizing video and passing it to the TPU, were performed on a desktop machine
and not on the more limited Raspberry Pi 5 to be used in the end product.
Third, using RGB images as input to these small networks quickly resulted in them overfitting to both color
and brightness.
Therefore, we limited the image input size to 512x512 pixels,
process grayscale images instead of RGB, and limited the number of U-Net sections to 3-4.
We evaluate our models in terms of categorical accuracy and cross-entropy.

\section{Results}
\subsection{Evaluation of Synthetic Segmentation Dataset}

We have collected terrain data in Iceland at 3 locations via photogrammetry,
and at 2 locations via LiDAR.
We also use two open-source photogrammetry datasets available through WebODM.
Table~\ref{table:datasets} lists the data sets used,
and Figure~\ref{figure:dataset_example} shows example pairs of images and masks from the soccer field dataset.
Notably, the data sets from WebODM tend to contain only orthogonal photography,
such that the models they produce appear realistic from a top-down view but less realistic from an angle,
especially on vertical surfaces.
We note additionally that the LiDAR reconstructions tend to capture smaller obstacles
and deeper cracks much more accurately
and make the surveys and subsequent data processing much quicker, on the scale of minutes to hours.

\subsection{Synthetic Evaluation of Terrain Classifiers}

We train a 4-stage U-Net with 512x512 resolution synthetic training and testing
images for a maximum of 100 epochs with early stopping
(monitoring the testing loss) with a patience of 15 epochs
and with a learning rate of \num{5e-5} with the Adam optimizer.
We use categorical accuracy and categorical cross-entropy as our accuracy and loss metrics, respectively.
The U-Net has four encoder steps, a bridge, and four decoder steps.
The encoder block pipeline has a 2D convolution, batch normalization, leaky ReLU, 2D convolution, batch normalization, leaky ReLU, and maxpool.
The decoder block pipeline has an upsampling layer, concatenation, 2D convolution, batch normalization, leaky ReLU, 2D convolution,
batch normalization, and leaky ReLU.
To conform with the available operations for the Google Coral,
we remove batch normalization layers,
replace upsampling with transpose convolution,
and replace each leaky ReLU layer with a PReLU layer with an untrainable alpha parameter.
We train in Keras~\cite{keras} and quantize classifiers destined for the Google Coral to a TFLite model.
This requires full quantization to unsigned 8-bit integers instead of the standard 32-bit float
and also requires a representative dataset to tune the network during the conversion.
The last step is to use the edge TPU compiler to convert the TFLite model into one compatible with the Google Coral.
For each experiment, we select a subset of the datasets in Table~\ref{table:datasets} for training
and testing.
We run each experiment 10 times, taking the model with the lowest testing loss.
This is feasible because the model is small, and the training time is 10--20 minutes.
The result is presented in Table~\ref{table:keras_experimental_results}.

\begin{table}[tb]
\caption{Best classifier accuracy, loss, and real-world validation accuracy.
The safety and danger thresholds for V2 are \(0.6\) and \(0.4\) respectively.}
\centering
\begin{tabular}{lrrrrrr} \toprule
Classifier & \multicolumn{2}{c}{Accuracy} & \multicolumn{2}{c}{Loss} & \multicolumn{2}{c}{Real-World} \\
            \cmidrule(lr){2-3} \cmidrule(lr){4-5} \cmidrule(lr){6-7}
           & Training      & Testing      & Training    & Testing    & ~~V1~~              & ~~V2~~           \\ \midrule
Best U-Net  & 0.667         & 0.815      & 0.613     & 0.373    & 0.778            & 0.833         \\
\bottomrule
\end{tabular}
\label{table:keras_experimental_results}
\end{table}

\subsection{Post-processing}
\label{section:post_processing}

\begin{figure}[t!]
        \centering

        \begin{subfigure}[t]{0.40\linewidth}
                \includegraphics[width=\linewidth]{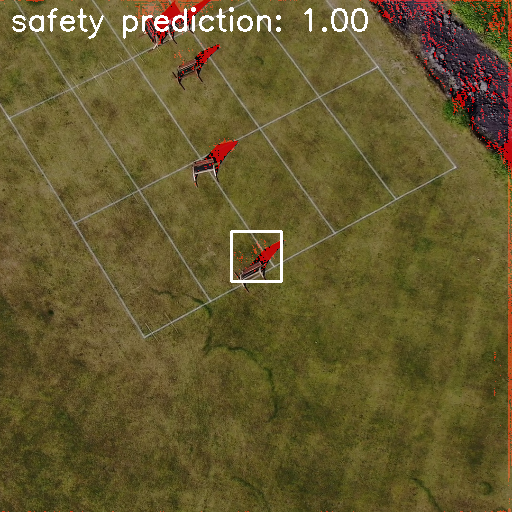}
                \caption{Raw output.}
                \label{figure:inference_raw}
        \end{subfigure}
        \begin{subfigure}[t]{0.40\linewidth}
                \includegraphics[width=\linewidth]{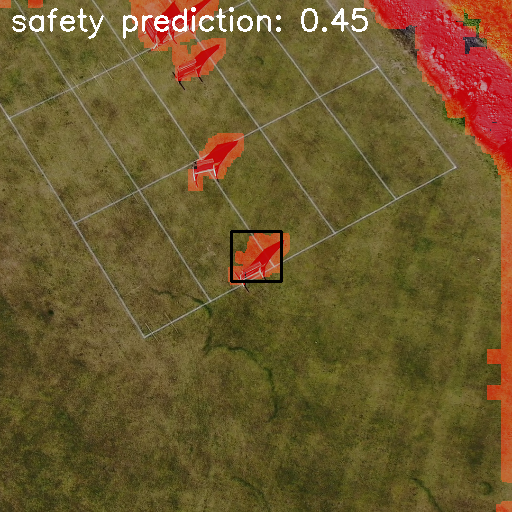}
                \caption{E1: box blur.}
                \label{figure:inference_e1}
        \end{subfigure}\\
        \begin{subfigure}[t]{0.40\linewidth}
                \includegraphics[width=\linewidth]{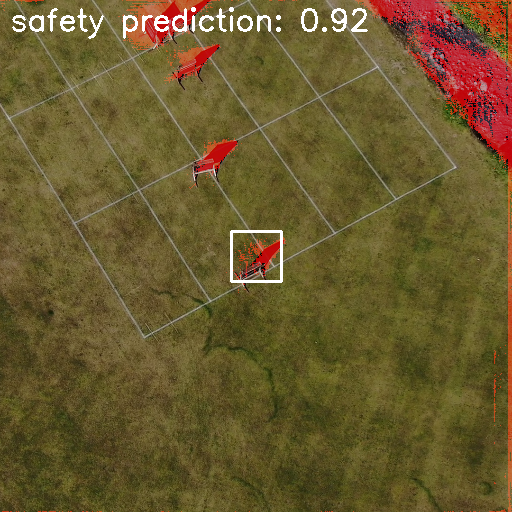}
                \caption{E2: temporal.}
                \label{figure:inference_e2}
        \end{subfigure}
        \begin{subfigure}[t]{0.40\linewidth}
                \includegraphics[width=\linewidth]{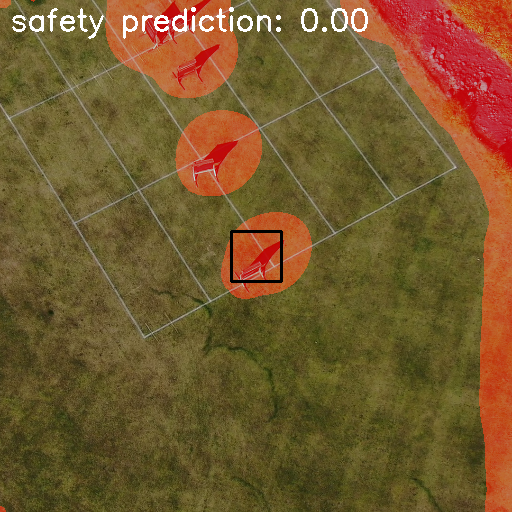}
                \caption{E1 \& E2.}
                \label{figure:inference_e1e2}
        \end{subfigure}        
        \caption{Example predictions with and without post-processing}
        \label{figure:post_processing}
\end{figure}

Figure~\ref{figure:inference_raw} shows an example prediction from the network
at the 10-second mark (the last frame)
of a video from the validation set.
The network outputs two masks: one for safety and one for danger.
For clarity, we only show the danger masks and other parts can be assumed safe.
Setting thresholds for converting these to a binary mask is yet another parameter;
we conduct a coarse parameter sweep, with the safety threshold $\theta_s$ being $[0.1, 0.2, 0.3, 0.4, 0.5, 0.6, 0.7, 0.8, 0.9]$
and the danger threshold being $1 - \theta_s$.
The central square region represents a candidate landing site that we want to evaluate
-- in this case, an archery target (unsafe for landing) --
and the evaluation is shown in the top left of the image.
As explained in Section~\ref{section:validation_dataset}, this evaluation is determined by whether the majority of
pixels in the square are considered safe or unsafe.
The square is white if the area is considered safe and black if it is considered unsafe.
Although many of the pixels in the box in Figure~\ref{figure:inference_raw} are red,
they are too sparse to reject the site -- this can be seen in the safety prediction of 1.00, indicating
that is has been deemed safe throughout the entire video.
In the spirit of erring on the side of caution, we would like to reject unsafe landing sites despite the
typical sparsity of unsafe classifications.
We thus propose two post-processing enhancements: E1 is a box blur with the job of patching holes between rejected
regions to create contiguous rejected regions,
and E2 is temporal smoothing, with the job of keeping track of temporally sparse unsafe classifications.
These are shown to reduce the safety prediction in Figure~\ref{figure:inference_e1} and Figure~\ref{figure:inference_e2}
respectively.
We experimented with values ranging from 1--10 frames for the temporal history and 7--19 for the box blur kernel size,
ultimately choosing a box blur kernel size of 15 for E1 and a temporal history of 5 frames for E2.
While E1 and E2 alone both reduce the safety prediction, only the combination of both completely reject the unsafe
landing site with a prediction of 0.00, as shown in Figure~\ref{figure:inference_e1e2}.

\subsection{Real-World Evaluation of Terrain Classifiers}

Using the landing site safety prediction method defined in Section~\ref{section:post_processing},
we conduct a secondary validation phase for the most promising classifiers.
These results are presented in the last two columns of Table~\ref{table:keras_experimental_results}
-- V1 represents the validation performance with safety and danger thresholds of 0.5, and V2 represents the validation performance
with thresholds tuned through a coarse parameter search.
We compare the network's aggregated prediction to our knowledge of the area, obtained by going to the area in person,
and determine the network's successful prediction rate over the validation locations in Figure~\ref{figure:data_collection_sites}.
Overall, the best U-net correctly classifies 15 of 18 of the 10-second validation videos,
i.e., it achieves an accuracy of 0.83, which we consider promising.
The classifiers can run at a rate of about 3.4 Hz on the drone payload,
which is fast enough for our task given that we can add filtering methods to make
the approach smooth.

There were some important trends in the real-world evaluation.
First, the classifiers almost universally classified a safe runway as unsafe and the unsafe, slanted gravel edge of a soccer
field as safe.
This seems to be a limitation of the fact that the classifier is appearance-based;
the straight, monochromatic white runway markings seem to appear as tall structures,
and the gravel does not change appearance significantly when slanted compared to when it is level (e.g., in a parking lot).
Further, orientation and altitude are important factors, and performance is unsurprisingly much better if these are within
the ranges of the synthetic training data, i.e.,
between 45 degrees below the horizon and vertically down, and at a distance of between 5 and 20 meters from the target.
For example, many classifiers correctly classified tall vegetation as unsafe when adequately close,
and as safe when farther away.
Finally, the small networks occasionally experience a ``burn-in'' where they produce persistently safe or unsafe
classifications in particular locations of the image, regardless of the content of the image.
This is most likely attributable to (1) disadvantageous initial conditions in the training process,
and 
(2) biases in the data set, e.g., the fact that edges of the images are often unsafe as a result of being outside the
generated terrain area, as shown in Figure~\ref{figure:dataset_example}.

In addition to lab tests, we conducted flights at multiple validation locations in the testing area,
with the method running onboard the drone.
The best way to showcase these experiments is with video:
\url{https://vimeo.com/j0shua/mmm2025-demo}.
The code for this project is available on Github~\cite{code_repo}.

\section{Conclusion \& Future Work}
We presented a pipeline for generating terrain classifiers to find landing sites for a drone autonomously
by analyzing video from the drone's RGB camera.
We automatically generated a synthetic dataset for image segmentation by reconstructing the terrain surveyed by the drone,
labeling the reconstructed terrain geometrically for landing safety,
and generating images and masks in simulation.
We used both photogrammetry and LiDAR datasets and used one openly-available photogrammetry dataset from WebODM to show the
method's flexibility.
We trained a U-Net on the synthetic dataset and
quantified its performance in the real world by performing inference on
10-second drone videos of 18 known-safe and known-unsafe validation sites.
The U-Nets correctly classified a maximum of 15 validation sites.
Finally, we ran the method in real-time onboard a drone equipped with a Raspberry Pi 5 and Google Coral TPU,
motivating our creation of a less than 1 MB network.

Future work should include collecting data from more real-world environments,
and generating more synthetic data sets for training.
More classifiers should be tested, e.g., auto-encoders, fully convolutional networks, and other segmentation methods
such as support vector machines -- with consideration of the constraints of embedded hardware so they can be embedded onboard
a drone.
It is also a point of interest to determine whether there is a difference in performance to be gained
by either photogrammetry or LiDAR over the other.
Finally, we will adapt this method so that it can generate commands to control the drone and execute the landings on its own,
similar to the method in~\cite{visir_landing}.

\bibliography{references}
\bibliographystyle{splncs04}

\end{document}